\documentclass[10pt,twocolumn,letterpaper]{article}

\usepackage{iccv}
\usepackage{times}
\usepackage{epsfig}
\usepackage{graphicx}
\usepackage{amsmath}
\usepackage{amssymb}

\usepackage[breaklinks=true,bookmarks=false]{hyperref}

\iccvfinalcopy 

\def\iccvPaperID{12} 

\ificcvfinal\fi

\begin{document}
\title{Synthetic Video Generation for Robust Hand Gesture Recognition in Augmented Reality Applications}

\newcommand*\samethanks[1][\value{footnote}]{\footnotemark[#1]}

\author{
Varun Jain\\
Carnegie Mellon University\\
\and
Shivam Aggarwal\samethanks \\
IIIT Delhi\\
\and
Suril Mehta \\
IIIT Delhi\\
\and
Ramya Hebbalaguppe \\
TCS Research, India \\
\and
{\tt\small  jainvarun@cmu.edu
, \{shivam16195, suril15104\}@iiitd.ac.in, ramya.hebbalaguppe@tcs.com
}
}
\maketitle
\ificcvfinal\thispagestyle{empty}\fi

\begin{abstract}
    Hand gestures are a natural means of interaction in Augmented Reality and Virtual Reality (AR/VR) applications. Recently, there has been an increased focus on removing the dependence of accurate hand gesture recognition on complex sensor setup found in expensive proprietary devices such as the Microsoft HoloLens, Daqri and Meta Glasses. Most such solutions either rely on multi-modal sensor data or deep neural networks that can benefit greatly from abundance of labelled data. Datasets are an integral part of any deep learning based research. They have been the principal reason for the substantial progress in this field, both, in terms of providing enough data for the training of these models, and, for benchmarking competing algorithms. However, it is becoming increasingly difficult to generate enough labelled data for complex tasks such as hand gesture recognition.
    The goal of this work is to introduce a framework capable of generating photo-realistic videos that have labelled hand bounding box and fingertip that can help in designing, training, and benchmarking models for hand-gesture recognition in AR/VR applications. We demonstrate the efficacy of our framework \footnote{\url{varunj.github.io/egogestvid}} in generating videos with diverse backgrounds.
\end{abstract}

\vspace*{-5mm}
\section{Introduction}

In the past, researchers have proposed deep neural network architectures consisting of ensemble of models that solve specific sub-tasks; For instance, sub-tasks such as hand candidate detection, fingertip detection and classification are used to achieve a larger goal of hand gesture recognition in first-person view~\cite{huang2016pointing}. For accurate hand gesture classification sans the depth data, each of the constituent models have to be trained separately and demand extensive human labour in annotating the data. The authors~\cite{huang2016pointing} use a manually annotated dataset containing over $90,000$ frames to introduce enough variability in background and lighting conditions so as to make the models robust. 

On the other hand, synthetically generated data is also being increasingly used of late to train and validate vision systems
~\cite{zimmermann2017learning}. This is especially true of areas in which obtaining huge amounts of data with ground truth is tedious. However, existing literature states that the performance of systems that are trained only on synthetic data is not at par with systems that are trained on real-world data due to the issue of domain shift~\cite{ros2016synthia}. This problem arises since the probability distribution over the parameters resulting from the process of generating the synthetic videos may diverge from the parameters that describe the real-world data. Divergence in critical parameters such as lighting, scene geometry, and camera parameters often lead to poor generalisability in models that are trained solely on synthetic data. 

Various works have derived or designed representations such as geometry and motion in synthetic domains that are quasi invariant to the problem of domain shift~\cite{baktashmotlagh2013unsupervised}. Ros et al.~\cite{ros2016synthia} have showed that augmenting large scale synthetic data with even a few real-world samples while training can relieve domain shift. Moreover, recent work in the field of generative adversarial learning~\cite{goodfellow2014generative, CycleGAN2017}, 
has shown how unlabelled samples from a target domain can be used to iteratively obtain better point estimates of parameters in generative models by minimising the difference between the generative and target distributions. 
Taking cues from the two ideas, we generate photo-realistic videos with different backgrounds and gesture patterns and hypothesise that given a large-scale dataset, one can design simpler frameworks that implicitly learn the global task of gesture recognition without needing to explicitly localise hands and fingertips.

\section{Proposed Framework}

\subsection{CycleGAN Based Approach}
We adapt the architecture for our generative networks from Zhu et al.~\cite{CycleGAN2017} who have shown impressive results for image-to-image translation. 

The network contains two $stride-2$ convolutions, two fractionally strided convolutions with $stride 1$, and several residual blocks. $9$ blocks are used for $256\times256$ size input images. To detect whether overlapping image patches are real or fake, the discriminator network uses $70\times70$ PatchGANs~\cite{isola2017image}. Such a patch-level discriminator architecture has fewer parameters than a full-image discriminator and can work on arbitrarily-sized images in a fully convolutional fashion.

\begin{figure}[h]
    \centering
    \includegraphics[width=1\linewidth]{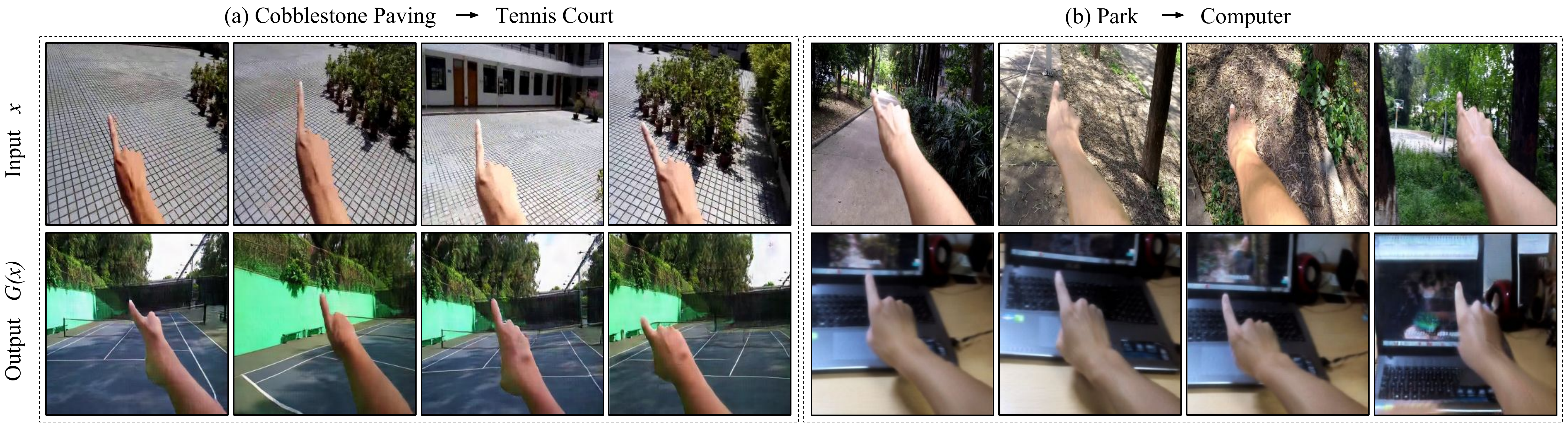}
    \caption{Results on two pairs of source and target domains. The upper row represents the real-world images and the lower row shows the synthesised image in the new domain. 
    }
    \label{fig:fig_inputoutput}
\end{figure} 
\subsection{Sequential Scene Generation with GAN}

\begin{figure}[h]
    \centering
    \includegraphics[width=1\linewidth]{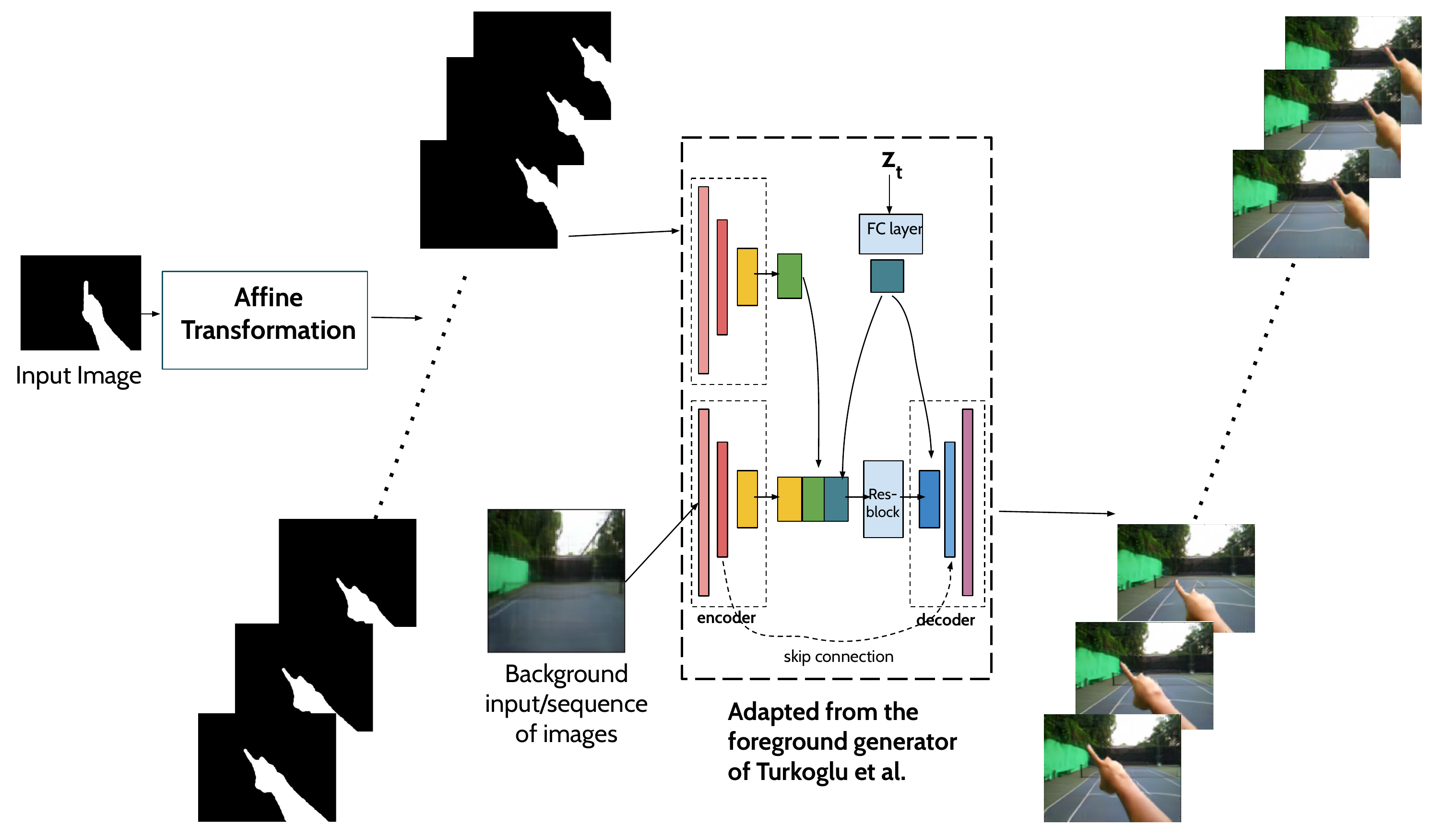}
    \caption{Proposed DNN: Given an input image, we apply gesture based affine transformation to generate a sequence of video frames. These masks passed in succession to the generator network results in a video sequence with given background image.}
\label{fig:fig_framework}
\end{figure}

We used the ability of the model outlined by Turkoglu et al.~\cite{SeqGAN} to generate video sequences with different backgrounds but same (or controlled) fingertip and hand as in the reference input image. The proposed framework sequentially composes a scene, breaking down the underlying problem into foreground and background separately. Our approach (figure~\ref{fig:fig_framework}) utilises the foreground generator as proposed by Turkoglu et al.~\cite{SeqGAN} to superimpose elements over the given background. 

\section{Experiments and Results}

\subsection{Experiment 1}
We use the Adam solver with a batch size of $1$. All models were trained from scratch with a learning rate of $0.0002$. The results were observed on varying number of epochs where the model was trained for $400$ epochs with the same learning rate and linearly decaying the learning rate over next $100$ epochs. The model was trained on a Tesla V100 GPU for $24$ hours.

We train our model on the SCUT-Ego-Finger dataset~\cite{huang2016pointing}. It has $93729$ manually annotated frames for hand detection and fingertip detection in first-person view. The dataset includes videos from $24$ different environments such as classroom, lake, canteen etc. We demonstrate our results in Figure~\ref{fig:fig_inputoutput} on two pairs of source and target domains: (a) $Cobblestone Paving \to Tennis Court$, and (b) $Park \to Computer$.

\subsection{Experiment 2}
\begin{figure}[h]
    \centering
    \includegraphics[width=0.9\linewidth]{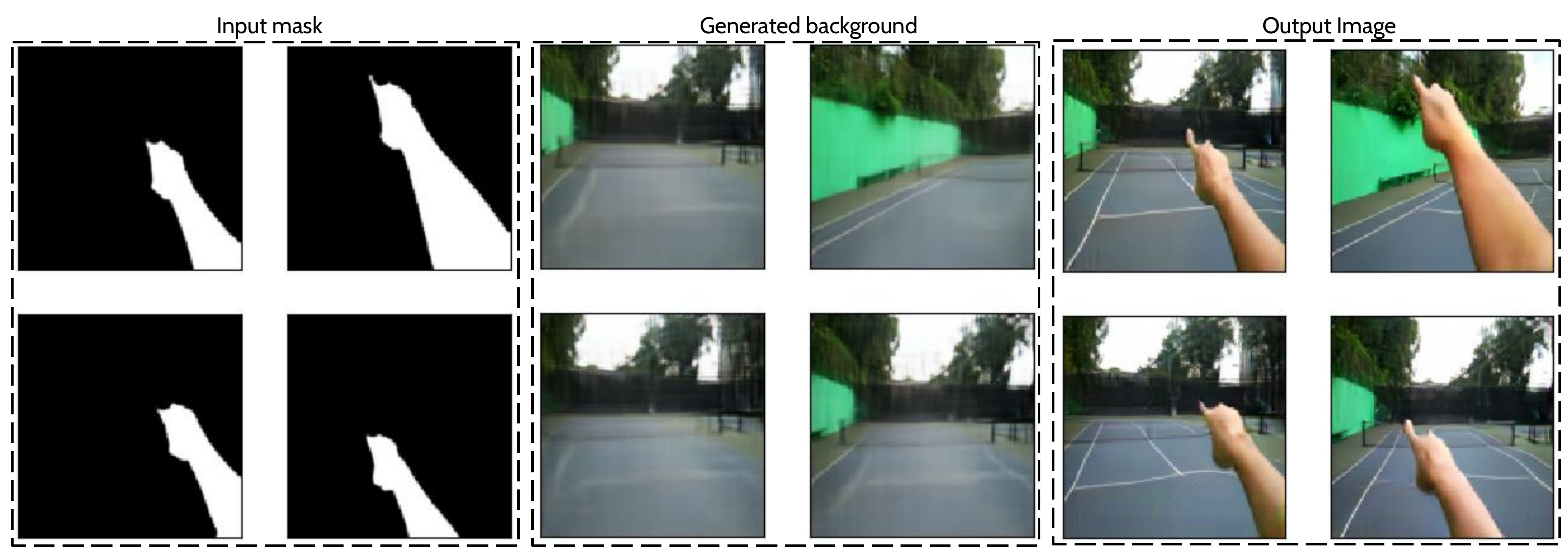}
    \caption{Results obtained using the proposed method. Given a segmentation map as input, the network generates a background image and an output image taking cues from the semantic layout map. The generated images are highly photo-realistic with little or no distortion.}
    \label{fig:fig_inp}
\end{figure}

We ran our experiments on a subset of the SCUT-Ego-Finger dataset~\cite{huang2016pointing}. Since we did not have ground-truth labelled semantic maps for our dataset, skin pixels are detected from the images using the skin-colour segmentation. We applied the GrabCut algorithm~\cite{GrabCut} for foreground extraction followed by skin-thresholding in HSV colour format. Morphological erosion is also applied to remove some of the isolated blobs.

We trained the foreground and background generator (for extracting background images from the dataset~\cite{huang2016pointing}) for 100 and 200 epochs respectively, with a batch size of 4. Figure~\ref{fig:fig_inp} demonstrates the complete use-case of the network. Because of the segmentation masks given as input to the model, the network is able to replicate hand and fingertip in the foreground fully.

\begin{figure}[h]
    \centering
    \includegraphics[width=0.8\linewidth]{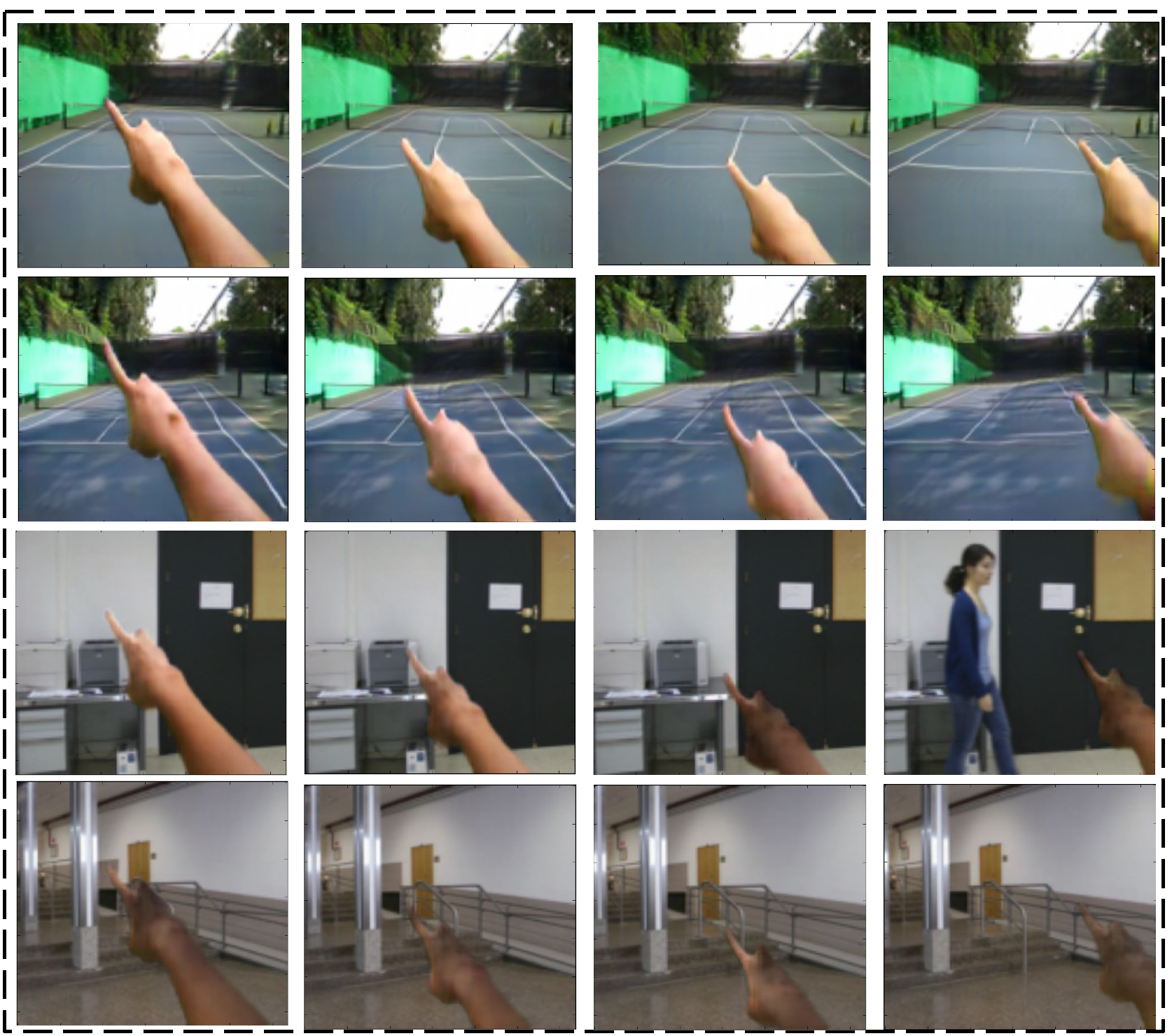}
    \caption{Results generated using same set of translated segmentation mask and fingertip location. Each row represents the synthesised images in the new domain.}
    \label{fig:fig_domain}
\end{figure}

Figure~\ref{fig:fig_domain} shows images with different background domains but the same mask layout as input. We observe that the synthesised images do not suffer from any artefacts as compared to images generated by CycleGAN~\cite{CycleGAN2017}. However, skin colour is a bit off in the fourth domain perhaps due to the 
texture of the background domain.

\begin{figure}[h]
    \centering
    \includegraphics[width=0.9\linewidth]{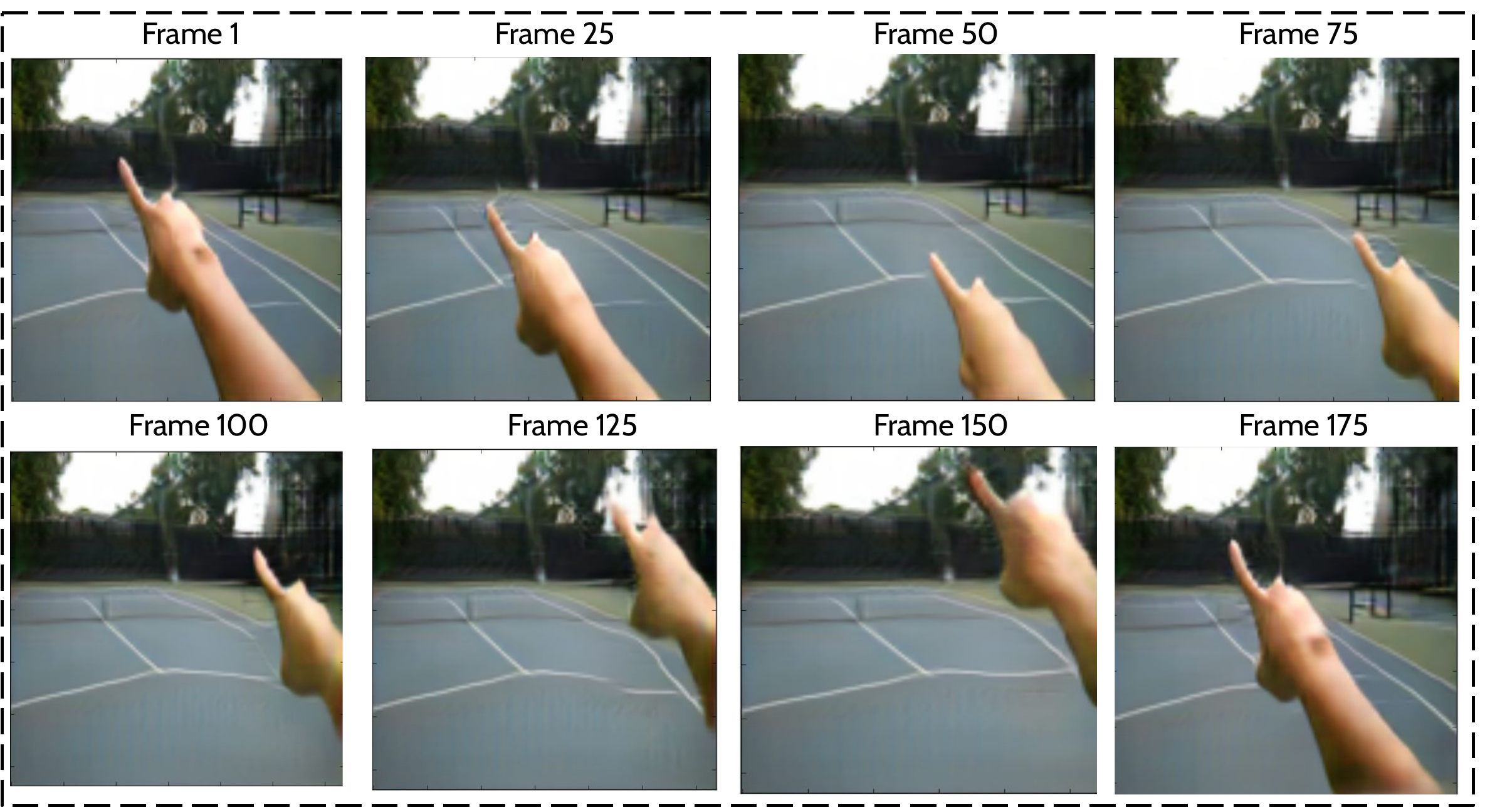}
    \caption{Generating a circle as an egocentric pointing gesture by applying orientation based mask affine transformation. Different frames depict gesture images as synthesised using the network. Note that complete gesture is obtained using a single layout mask as reference.}
    \label{fig:fig_circle}
\end{figure}

We extend this idea to generate egocentric gestures such as fingertip going down, up, left, and right. One such example has been demonstrated in Figure~\ref{fig:fig_circle}.

\section{Future Work}
The realisation of our end goal of generating photo-realistic videos with enough variability in background, lighting, and other such parameters that can help in designing, training, and benchmarking models for hand-gesture recognition 
would involve designing a model that introduces variations in the background features and some features present on the hand. We would like to experiment the inclusion of a recurrent network into the current framework which could 
generate photo-realistic hand movements corresponding to any given spatio-temporal sequence corresponding to an arbitrary input gesture. Finally, we observe that the background might change suddenly between consecutive frames leading to a jittery video and we would like to experiment with ways to make the background coherent across frames.

\section{Conclusion}

We have demonstrated a network capable of synthesising photo-realistic videos and show its efficacy by generating videos of hand gestures. We believe that this would help in the creation of large-scale annotated datasets, which, in turn, would encourage the development of novel neural network architectures that can recognise hand gestures from single RGB streams without the need of specialised hardware such as multiple cameras and depth sensors.

{\small
\bibliographystyle{ieee_fullname}
\bibliography{egpaper_final}
}

\end{document}